\documentclass{article}
\usepackage{arxiv}
\usepackage[utf8]{inputenc} 
\usepackage[T1]{fontenc}    
\usepackage{hyperref}       
\usepackage{url}            
\usepackage{booktabs}       
\usepackage{amsfonts}       
\usepackage{nicefrac}       
\usepackage{microtype}      
\usepackage{lipsum}		
\usepackage{graphicx}
\usepackage{natbib}
\usepackage{doi}
\usepackage{multicol}
\usepackage{pgfplots}
\usepackage{array}

\pgfplotsset{compat=1.7}

\title{Benchmarking of Deep Learning models on 2D Laminar Flow behind Cylinder}
\date{}
 \author{
    Mritunjay Musale\\
    Department of Computer Engineering,\\
    K. J. Somaiya College of Engineering, Somaiya University\\
	  Mumbai, India 400077 \\
  \texttt{mritunjay.m@somaiya.edu}
  \And
  Vaibhav Vasani\\
  Department of Computer Engineering,\\
  K. J. Somaiya College of Engineering, Somaiya University\\
  Mumbai, India 400077 \\
  \texttt{vaibhav.vasani@gmail.com}
 }
 
\begin{document}
  \maketitle

\begin{abstract}
  The rapidly advancing field of Fluid Mechanics has recently employed Deep Learning to solve various problems within that field.
In that same spirit we try to perform Direct Numerical Simulation(DNS) which is one of the tasks in Computational Fluid Dynamics, using three fundamental architectures in the field of Deep Learning that were each used to solve various high dimensional problems.
We train these three models in an autoencoder manner, for this the dataset is treated like sequential frames given to the model as input.
We observe that recently introduced architecture called Transformer significantly outperforms its counterparts on the selected dataset. 
Furthermore, we conclude that using Transformers for doing DNS in the field of CFD is an interesting research area worth exploring.

\end{abstract} 

\section{Introduction}
\paragraph{}
Recently the advancements in Deep Learning(DL)\citep{menghani2021efficient} has led to a revolution in various domains of science and engineering\citep{brunton_kutz_2019,razzak2018deep,kamath2019deep,raghu2020survey}.
One of the key areas where DL has shown significant progress in terms of research and application is in computer vision.
Convolutional Neural Networks (CNNs)\citep{LeCun1998GradientbasedLA} has been a key enabler for such progress.
CNNs have been the backbone of many architectures that have led to the advancements of DL for computer vision relates tasks such as image synthesis\citep{goodfellow2014generative,Karras2021AliasFreeGA}, video masking\citep{Lu2021OmnimatteAO} and several other images and video processing tasks\citep{9563948,8917633}.
DL has also made huge strides in the field of Natural Language Processing (NLP) with the introduction of the transformer\citep{vaswani2017attention} architecture and its variants\citep{Devlin2019BERTPO,liu2019roberta,Brown2020LanguageMA}.

\paragraph{}
Text data which is usually one-dimensional feature vector of arbitrary length and images are 2 dimensional or 3 dimensional data which are hard to train using a simple fully-connected neuron based models.
CNNs and transformers have allowed DL to tackle problems dealing with high number of dimensions\citep{georgiou2020survey}. 
This coupled with advancements in hardware\citep{8998988} has led to researchers making deeper and wider models for various tasks which weren't possible before.
One such domain which contains high dimensional and high amount of data is the field of Computational Fluid Dynamics.

\paragraph{}
CFD is a field that mostly consists of numerical solutions of differential equations governing mass, momentum and energy with respect to fluids\citep{date2005introduction}. 
Experiments in CFD generate huge amounts of data after each iteration and in order to gain insights from such data traditionally advance algorithms, data mining, compression, databases of flows\citep{10.1145/1362622.1362654} and several other approaches have been used.
Recently there has been a trend in using DL to solve various tasks in CFD.
Deep Learning models can be trained in linear time due advancements in software and hardware, which has been a key motivator for CFD being applied using DL.
CFD are extremely complex and require a lot of computation for calculating fluid flows and are often bound by the amount CPU resources of a system, whereas in contrast DL can be applied to various types of hardware\citep{8998988}, and we can get huge improvements in performance. 
Large availability of data, coupled with DL being optimized for various hardware has led to a growing research in ways to make CFD tasks solvable using DL.
We discuss various approaches where DL has been utilized for CFD in the next section\ref{related_work}.

\section{Related Work}
\label{related_work}

Deep Learning has been previously used for acceleration of various simulations in the past, some of those works are discussed in this section.

\paragraph{Machine learning–accelerated computational fluid dynamics}
\citep{doi:10.1073/pnas.2101784118} has applied Deep Learning in an upscaler fashion using super resolution approach.
Wherein the authors first generate 2D flow fields in a coarse grid of \emph{64 x 64} and upscale it to \emph{2048 x 2048}. 
They demonstrate an \emph{86x} speed up as compared to a Direct Numerical Simulation (DNS) which solves at the full \emph{2048 x 2048} resolution, while achieving competitive results.
They show that generating coarse flows at a low resolution and then running it through the DL based upscaler takes far less resources than doing so natively.
This paper doesn't necessarily do the computation of CFD by itself but much rather uses an existing CFD output and upscales it while filling in the missing information using a Deep Learning model.

\paragraph{Reynolds averaged turbulence modelling using deep neural networks with embedded invariance}
\citep{ling_kurzawski_templeton_2016} introduce a special tensor basis layer which is responsible for capturing information irrespective of the coordinate frame of the observer, which they call it as rotational invariance.
They use a standard MLP based model in conjunction with their custom tensor basis layer, for predicting the Reynolds stress anisotropy values.  
They show that embedding their tensor basis layer performed significantly better than and was able to give more accurate prediction of the stress values than simple MLP based model.
Furthermore, they also conduct an interpolation test where their model outperforms the conventional RANS models in the two test cases designed by them.

\paragraph{Deep Neural Networks for Data-Driven Turbulence Models}
\citep{beck2018deep} uses Deep Neural Networks (DNNs) for Large Eddy Simulations (LES), where only the largest turbulent scales are resolved, and smaller ones are modelled.
The authors use a Convolutional Neural Network (CNN) for demonstrating LES including the closure terms.
They also found out that ResNets\citep{he2016deep} depending upon the coarse input can predict the closure terms with good accuracy.    
They also point out that, the model's performance was hindered by the amount of data that was available to them and not the architecture itself.

\paragraph{Deep Learning Methods for Reynolds-Averaged Navier-Stokes Simulations of Airfoil Flows}
\citep{Thuerey2018DeepLM} propose a U-Net\citep{ronneberger2015u} based approach for predicting Reynolds Flow Field. 
The U-Net model receives three constant fields as input which contains the airfoil shape, and it predicts what the Reynolds Flow Field for that shape would look like.
The authors benchmark their model on two different Reynolds numbers and angle of attacks, and conclude that generalization performance not only depends on type and amount of training data, but also on the representative capacities of the chosen CNN architecture.

\paragraph{Shallow Neural Networks for Fluid Flow Reconstruction with Limited Sensors}
\citep{erichson2020shallow} use a shallow neural network, where they reconstruct the original fluid flow using limited sensor information.
They use a 3 layered architecture consisting purely of linear layers with ReLU activation for their experiments. 
They show that using a shallow decoder network with limited sensor information yields significantly better results as compared to a proper orthogonal decomposition (POD) or principal component analysis (PCA) based approach.
They also show that the model able to show much better results on interpolation tasks as compared to extrapolation where the model has to predict the future frames which it hasn't seen before on a turbulence flow dataset.

\section{Methodology}
\begin{figure}
  \centering
\begin{tabular}{c}
  \begin{minipage}{0.6\linewidth}
    \label{FlowAutoencoder_architecture}
    \includegraphics[width = \linewidth]{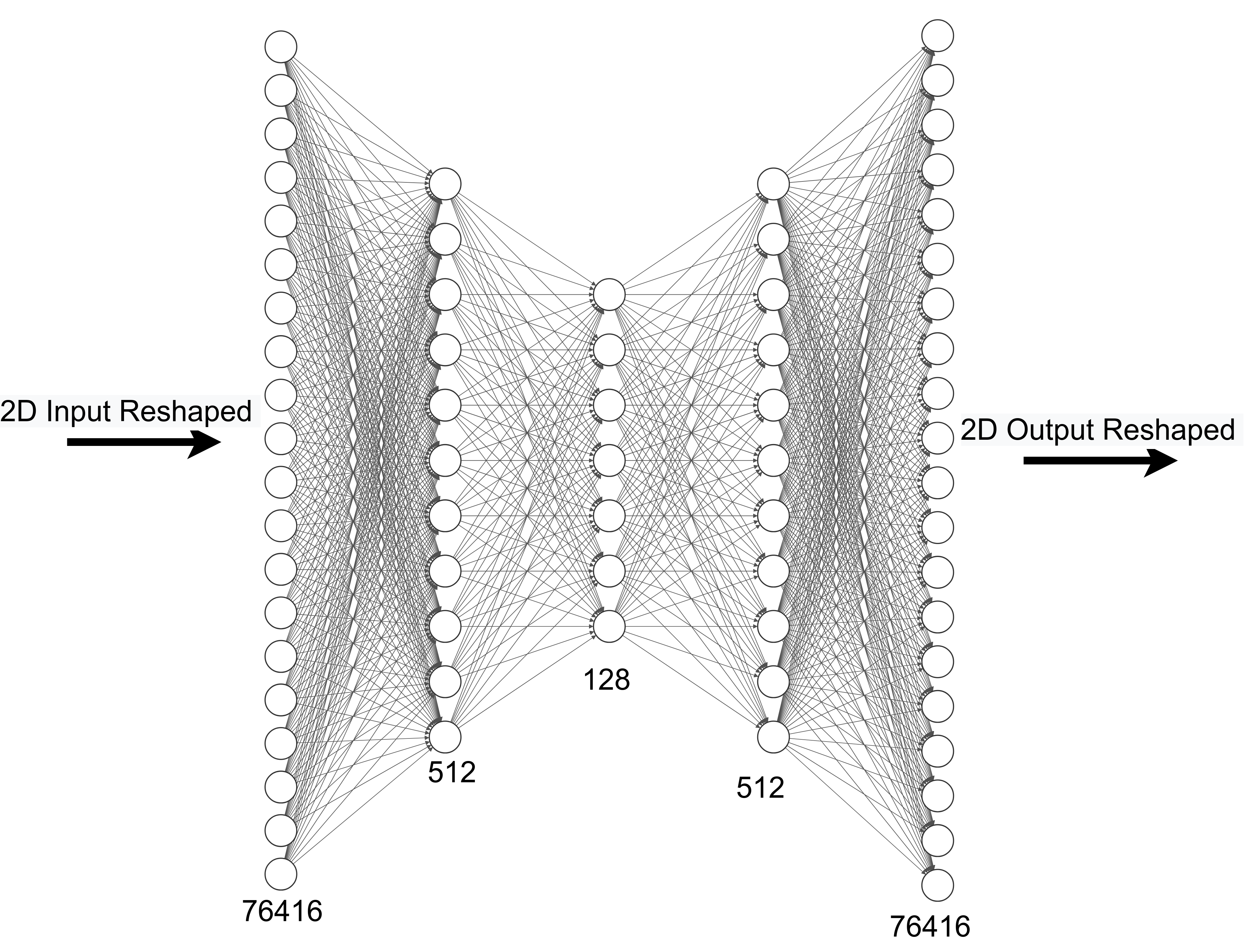}
    \caption{Architecture of FlowAutoencoder }
  \end{minipage}
  \\
  \hline
  \begin{minipage}{\linewidth}
    \label{FlowConvAutoencoder_architecture}
    \includegraphics[width = \textwidth]{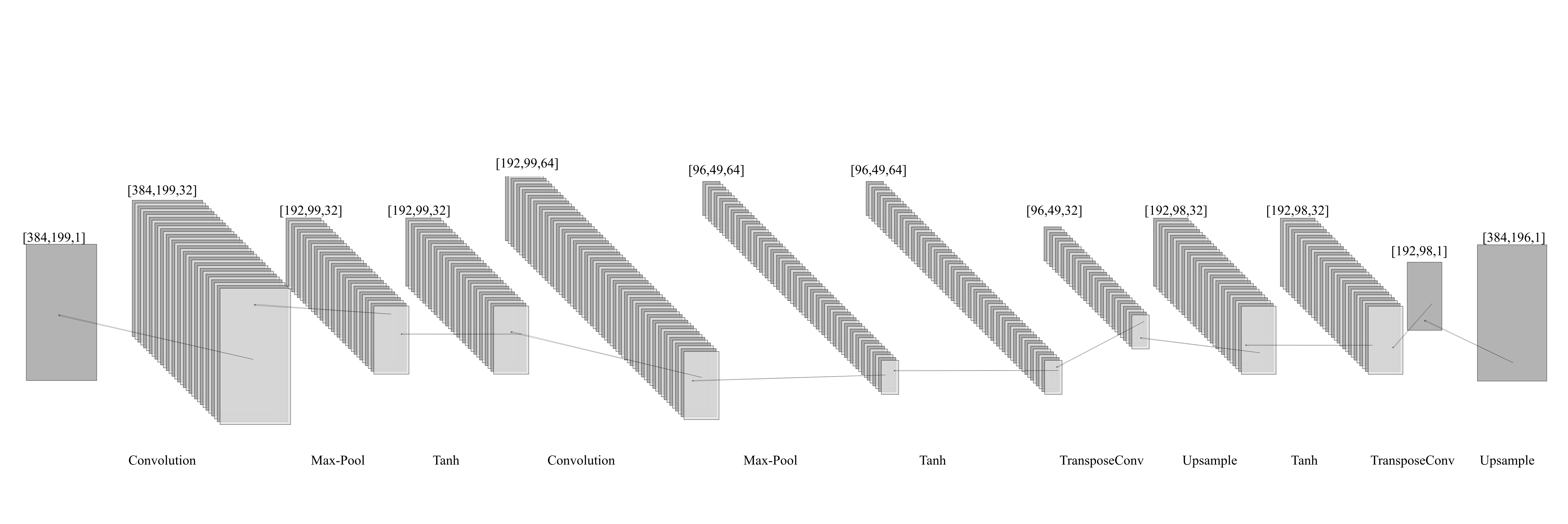}
    \caption{Architecture of FlowConvAutoencoder }
  \end{minipage}
  \\
  \hline
  \\
  \begin{minipage}{\linewidth}
    \label{FlowTransformer_architecture}
    \includegraphics[width = \linewidth]{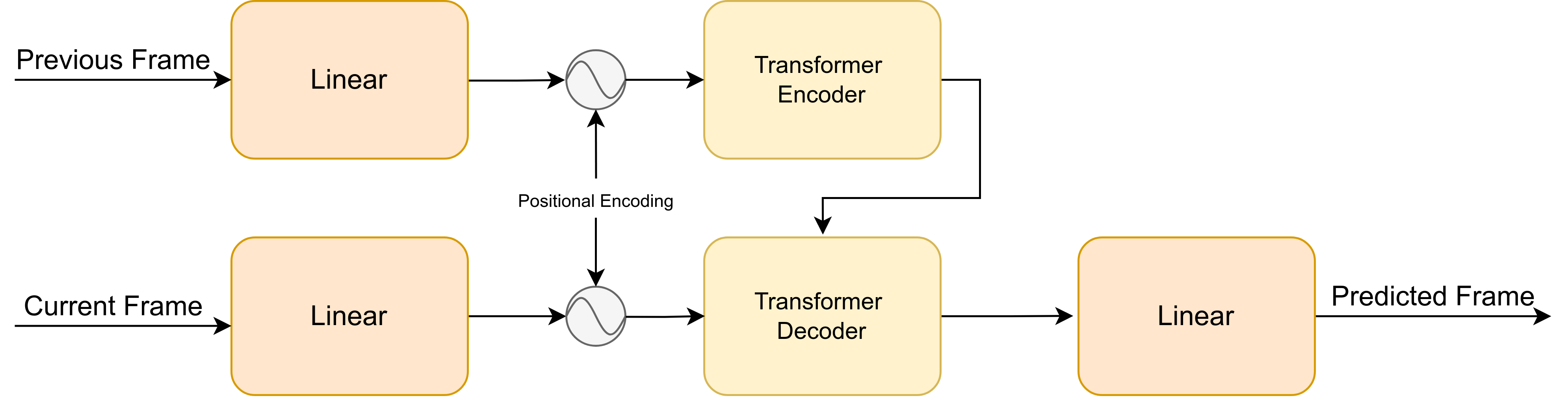}
    \caption{Architecture of FlowTransformer }
  \end{minipage}\\
\end{tabular}  
\end{figure}

\subsection{Overall view}
Previous approaches have used Deep Learning as an upscaler\citep{doi:10.1073/pnas.2101784118} or as a black box approach with custom layer for calculating Reynolds stress values\citep{ling_kurzawski_templeton_2016} or as a way to predict what the flow fields looked like from limited sensor information\citep{erichson2020shallow}.
We benchmark Deep Learning model to perform DNS directly, for this we treat each time step of the simulation as a 2D image consisting of the raw values of the flow simulation at that time step.
We feed the current frame as input to each of the 3 models (\ref{FlowAutoEocder_paragraph}\ref{FlowConvAutoEncoder_paragraph},\ref{FlowTransformer_paragraph} gets additional frame explained below) 
then these models will predict the next frame.
For this we propose three different models which are trained in an autoencoder manner\citep{bank2020autoencoders} - 
a simple autoencoder based on linear layers, a fully convolutional autoencoder and a transformer based model. 
Autoencoders expect their input and their output to be of the same dimensions, which works for our setup since both our inputs and outputs are images of the same dimensions.
Each of our proposed models is discussed in the next few sections, followed by the dataset and training configuration of the setup.

\subsection{FlowAutoEncoder}
\label{FlowAutoEocder_paragraph}
The FlowAutoEncoder model consists of three fully connected layers\ref{FlowAutoencoder_architecture}.
Each layer in the network consists of 512, 128 and 512 features with Tanh activation respectively\citep{nwankpa2018activation}. 
The two-dimensional input is flattened to a vector of length 512 and given as input to the first layer.
The vector dimensions gets reduced in 2nd layer to 128 and gets converted back to a vector of 512 in the last payer.
The prediction of the model is a vector of length 512, which is then reshaped to a 2D image.

\subsection{FlowConvAutoEncoder}
\label{FlowConvAutoEncoder_paragraph}
FlowConvAutoencoder is a fully convolutional autoencoder\ref{FlowConvAutoencoder_architecture}.
The fully convolutional nature of this auto encoder allows us to treat the input as an image and not as a feature vector.  
The encoder consists of two convolutional + max pooling + tanh layers which will reduce the dimensions of the input, followed by a decoder that does convolutional + upscale + tanh will restore the dimensions to the original input size.
Doing feature reduction and expansion will make the model learn features that will be useful for the prediction of the future frame. 
The output of the decoder is padded to make it same as the dimensions of the input.
In terms of activation function we selected Tanh, since the values from the simulation frame range from negative to positive and ReLU doesn't allow negative values to pass\citep{nwankpa2018activation} which is undesirable for us.

\subsection{FlowTransformer}
\label{FlowTransformer_paragraph}
The FlowTransformer is based on the transformer architecture described by~\citep{vaswani2017attention}. 
We use the transformer implementation provided with the PyTorch library~\citep{NEURIPS2019_9015}. 
The transformer consists 1 encoder, 1 decoder and 8 multiheadattention heads respectively we call our transformer based model as FlowTransformer\ref{FlowTransformer_architecture}.
The previous frame is given to the encoder and the current frame is given to the decoder as input, we get the predicted future frame in the output.
For FlowTransformer we flatten the 2D input and give to a linear layer of length 512, we do this for previous frame, current frame and to the output from the transformer.
We add positional embeddings to both feature vectors from linear layers of the input before giving to the transformer as described in~\citep{vaswani2017attention}.
The prediction vector of length 512, which goes through a linear layer is then reshaped to a 2D image which corresponds to the original input size.

\subsection{Dataset}
The selected dataset for this paper is consists of 2D laminar flow data from behind a cylinder~\citep{erichson2020shallow}.
We treat the problem of 2D laminar flow behind a cylinder as a sequence based task.
Each frame is a 2D image, and the sequence is a sequence of frames.
The dataset consists of 151 flow frames, form which 120 are used for training and 31 for testing.
Each model gets current frame as input and predicts the next frame, whereas the FlowTransformer additionally gets the previous frame as input along with the current frame.

\subsection{Training Configuration}
\paragraph{}
For the given dataset we train all the models for $N$ frames. 
Where $(N*0.8)$ sequential frames are used for training and $(N*0.2)$ sequential frames are used for testing.
We use a batch size of $12$, and we train for  $20$ epochs.
We optimize the model using the Adam optimizer~\citep{kingma2014adam} with a learning rate of $0.001$.
For evaluation of the model, we use Peak Signal to Noise Ratio (PSNR) and Structural Similarity Index Measure (SSIM) as the metrics, along with Mean Squared Error (MSE) as the objective function:
\begin{equation}
\mathcal{MSE}(Y,\hat{Y}) = \frac{1}{N} \sum_{i=1}^N (Y_i - \hat{Y}_i)^2
\end{equation}

\section{Results and Discussion}
\begin{figure}[!htp]
  \label{fig:results_figure}
  \centering
  \begin{tabular}{c c c}
    
    \begin{tikzpicture}[scale=0.7]
      
      \begin{axis}[
        xlabel={Training steps},
        ylabel={Loss},
        enlargelimits=false,
        grid=both,
        scale only axis=true,
        width={0.36\textwidth},
        ]
        \addplot+[no markers] table[x=step,y=FlowTransformerValues,col sep=comma] {plotting_data/Train_step_loss.csv}; 
        \addplot+[no markers] table[x=step,y=FlowConvValues,col sep=comma] {plotting_data/Train_step_loss.csv}; 
        \addplot+[no markers] table[x=step,y=FlowAutoencoderValues,col sep=comma] {plotting_data/Train_step_loss.csv}; 
        \addlegendentry{FlowTransformer}
        \addlegendentry{FlowConvAutoEncoder}
        \addlegendentry{FlowAutoEncoder}
        
      \end{axis}
    \end{tikzpicture}

    \begin{tikzpicture}[scale=0.7]
      
      \begin{axis}[
      xlabel={Training steps},
      ylabel={PSNR},
      enlargelimits=false,
      grid=both,
      scale only axis=true,
      width={0.36\textwidth},
      legend pos=south east,
      ]
      \addplot+[no markers] table[x=step,y=FlowTransformerValues,col sep=comma] {plotting_data/PSNR_training.csv}; 
      \addplot+[no markers] table[x=step,y=FlowConvValues,col sep=comma] {plotting_data/PSNR_training.csv}; 
      \addplot+[no markers] table[x=step,y=FlowAutoencoderValues,col sep=comma] {plotting_data/PSNR_training.csv}; 
      \addlegendentry{FlowTransformer}
      \addlegendentry{FlowConvAutoEncoder}
      \addlegendentry{FlowAutoEncoder}

    \end{axis}
  \end{tikzpicture}

  \begin{tikzpicture}[scale=0.7]
    
    \begin{axis}[
      xlabel={Training steps},
      ylabel={SSIM},
      enlargelimits=false,
      grid=both,
      scale only axis=true,
      width={0.36\textwidth},
      legend pos=south west,
      ]
      \addplot+[no markers] table[x=step,y=FlowTransformerValues,col sep=comma] {plotting_data/SSIM_training.csv}; 
      \addplot+[no markers] table[x=step,y=FlowConvValues,col sep=comma] {plotting_data/SSIM_training.csv}; 
      \addplot+[no markers] table[x=step,y=FlowAutoencoderValues,col sep=comma] {plotting_data/SSIM_training.csv}; 
      \addlegendentry{FlowTransformer}
      \addlegendentry{FlowConvAutoEncoder}
      \addlegendentry{FlowAutoEncoder}

    \end{axis}
  \end{tikzpicture}

\end{tabular}

\caption{Evaluation of metrics during training: MSELoss(left), PSNR(center) and SSIM(right).}

\end{figure}
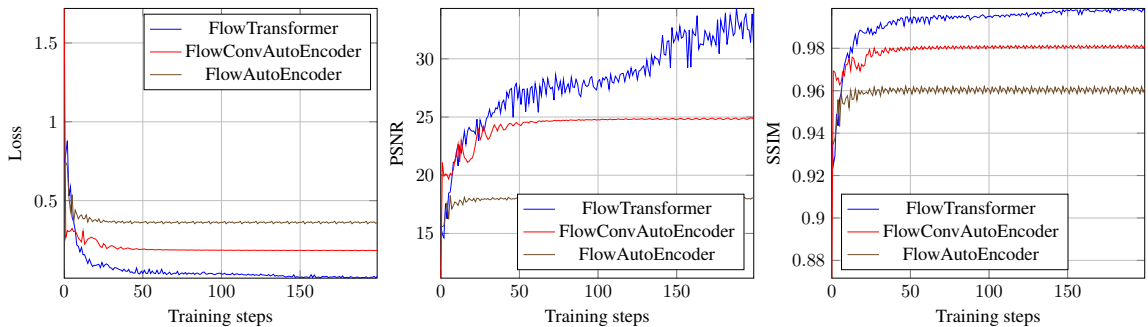

\paragraph{}
We evaluate the model after each training step as shown in figure\ref{fig:results_figure}, based on the selected metrics.
FlowAutoencoder, Flow ConvAutoencoder and FlowTransformer were able to achieve $0.361, 0.1846$ and $0.01952$ loss respectively on their last training step.
In terms of PSNR, FlowAutoencoder, Flow ConvAutoencoder and FlowTransformer achieve $18.0038, 24.8092$ and $30.8653$ PSNR during their last step of training.
Similarly, all three models achieve $0.9612, 0.9812$ and $0.9974$ SSIM respectively on their last step of training.
In figure\ref{fig:results_figure} we can see that the FlowTransformer performs the better than the other two models in all the three metrics.
The FlowTransformer is also able to better represent the values within the flows as shown in \ref{appendix_sample_outputs}.

\begin{table}[!htp]
  \centering
  \scriptsize
  \resizebox{0.6\textwidth}{!}
  {
  \begin{tabular}{lrrrr}\toprule

    \textbf{Model} & \textbf{Test Loss} & \textbf{PSNR} & \textbf{SSIM} \\
    \midrule
    \textbf{FlowAutoencoder }&0.3391 &18.3059 &0.9650 \\
    \textbf{FlowConvAutoencoder} &0.1828 &24.8892 &0.9818 \\
    \textbf{FlowTransformer }& \textbf{0.0094} & \textbf{33.7320} & \textbf{0.9985} \\
    \bottomrule
  \end{tabular}
  }
  \caption{Achieved test results }\label{tab: }
  \end{table}

\paragraph{}
We observe a similar trend during testing. 
FlowTransformer achieves significantly less loss than the other models during testing.
The FlowTransformer also shows much better PSNR and SSIM than the other models.
\cite{LeCunYoutube} describes one possible explanation as to why simple linear neural networks and Convolutional Neural Networks are not able to predict the future in terms of next frame prediction task.
We believe the same phenomenon is also present in the case of the FlowAutoEncoder model, where it produces the average values of the two flows that flow in alternating fashion.
In contrast, FlowTransformer is able to capture the dynamic natural of the problem and produce a flow field that much closer to that of the ground truth. 
The Self-Attention mechanism\citep{vaswani2017attention} must have played a key role in allowing the FlowTransformer model to outperform other models by means of allowing the model to adapt as per the stochastic nature of the input.
We show the sample outputs observed during training in appendix \ref{appendix_sample_outputs}.

\section{Conclusion}
\paragraph{}
We benchmarked, three different generations of models used in Deep Learning over the years, which were benchmarked on 2D Laminar Flow\citep{erichson2020shallow}.
We showed how the FlowTransformer after given the previous and current flow, can predict the next flow.
Due to resource constraints the dataset and models we used were significantly small as compared to other datasets available for deep learning\citep{5206848} and as compared to larger models like \citep{dosovitskiy2020image} or \citep{bao2021beit}.
We believe with a much larger dataset and models, the models can be better at the task of predicting the fluid flow.

\newpage
\bibliographystyle{unsrtnat}
\bibliography{bibliography}

\newpage
\appendix
\section{Sample outputs}
\label{appendix_sample_outputs}
\begin{figure}[htp]
  \centering
  
    \begin{tabular}
      {@{}ccccc@{}}
      \toprule
        \textbf{Model} & 
        \textbf{FlowAutoencoder }&
        \textbf{FlowConvAutoencoder} &
        \textbf{FlowTransformer }&
        \\
        \midrule
        
        \textbf{1\textsuperscript{st} step} &

        \includegraphics[scale=0.28]{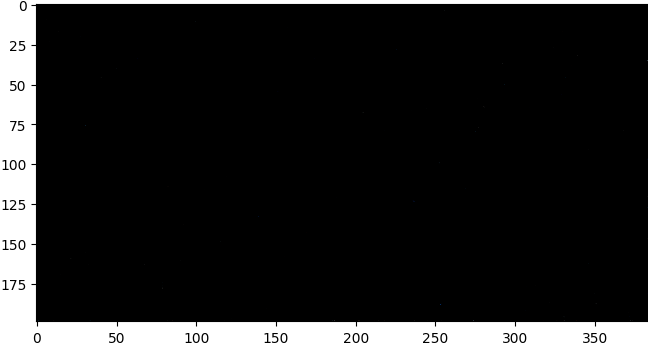} &
        \includegraphics[scale=0.28]{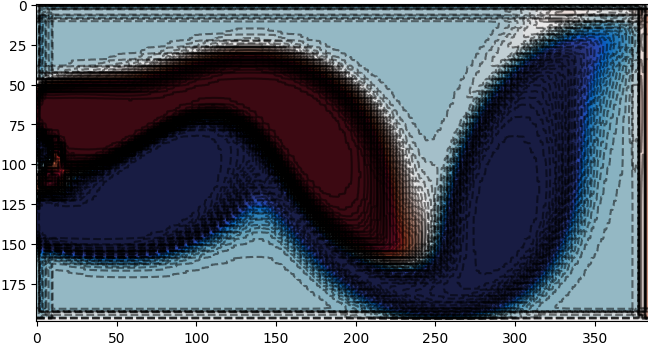} &
        \includegraphics[scale=0.28]{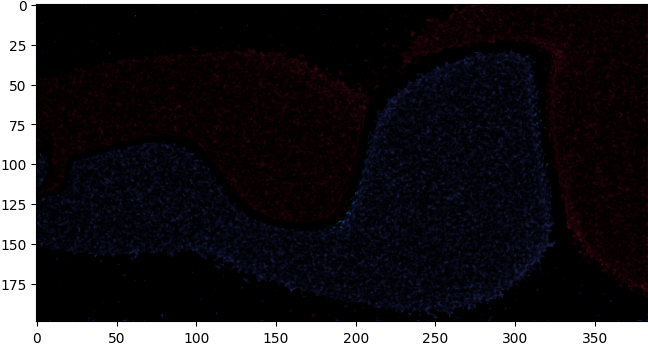} \\

        \textbf{100\textsuperscript{th} step} &
        
        \includegraphics[scale=0.28]{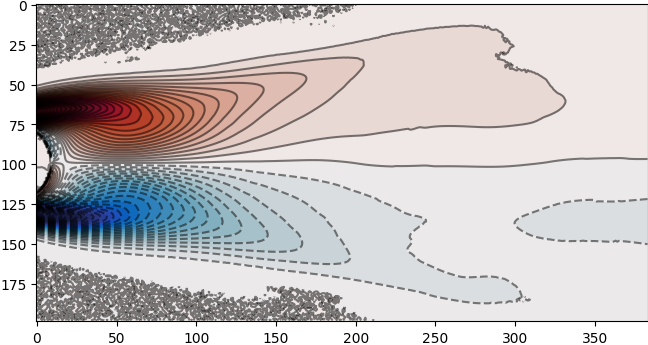} &
        \includegraphics[scale=0.28]{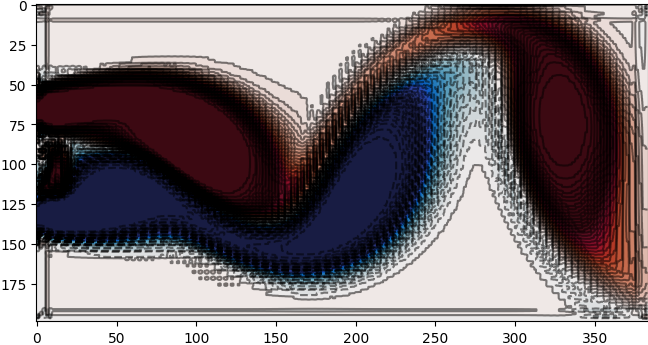} &
        \includegraphics[scale=0.28]{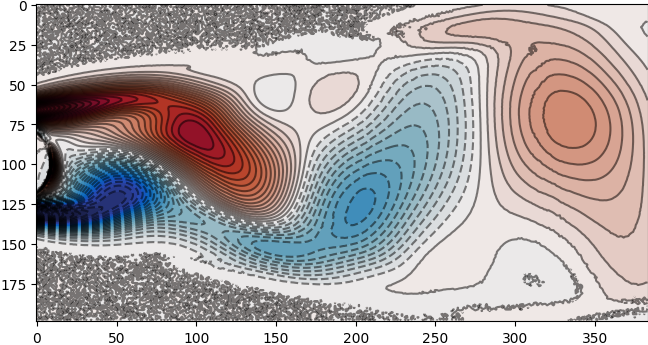} \\

        \textbf{200\textsuperscript{th} step}  &
        
        \includegraphics[scale=0.28]{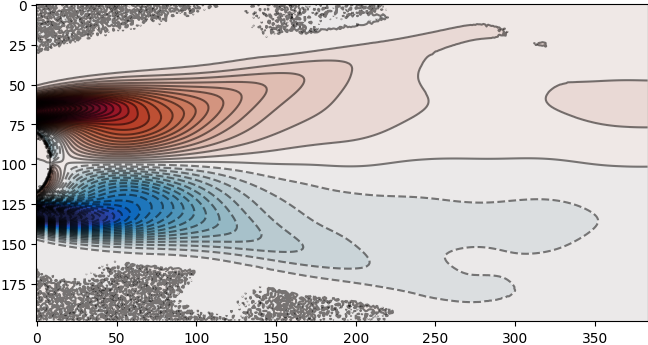} &
        \includegraphics[scale=0.28]{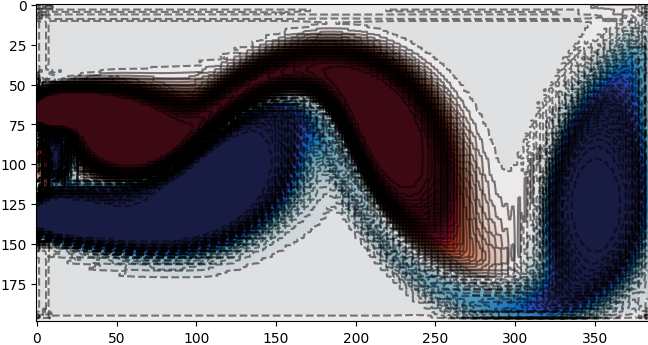} &
        \includegraphics[scale=0.28]{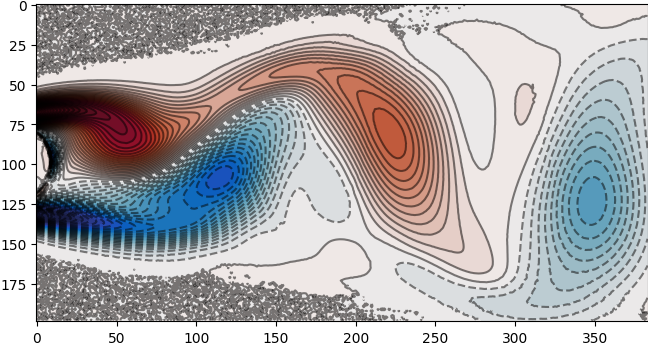} \\
        
        \textbf{Ground truth} \\(at 200\textsuperscript{th} step) &
        \includegraphics[scale=0.28]{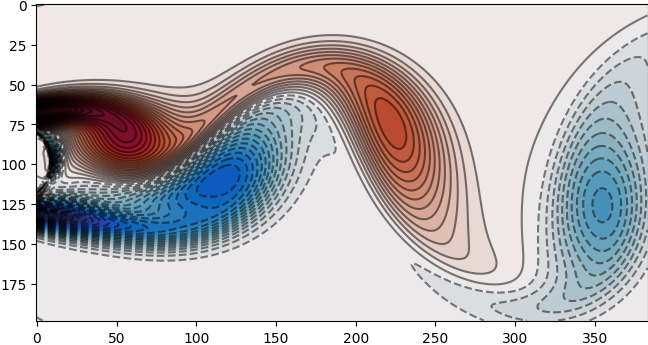} &
        \includegraphics[scale=0.28]{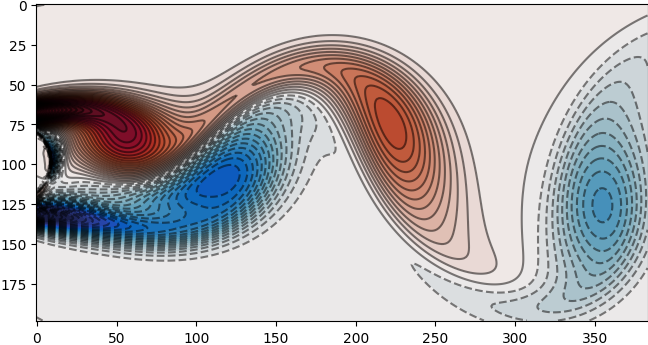} &
        \includegraphics[scale=0.28]{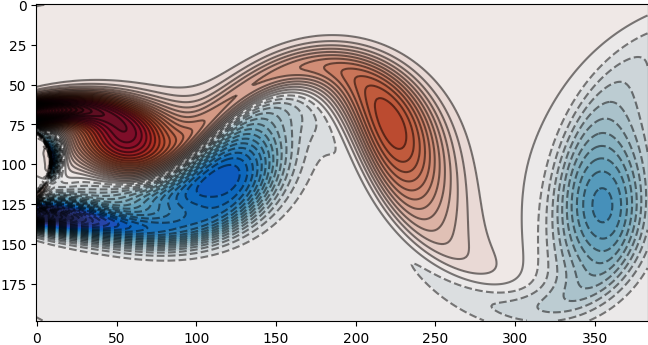} \\
      \bottomrule 
    \end{tabular}
 
    \caption{Sample outputs } 
    
\end{figure}


\end{document}